\documentclass[
]{ceurart}

\begin{document}

\copyrightyear{2021}
\copyrightclause{Copyright for this paper by its authors.
  Use permitted under Creative Commons License Attribution 4.0
  International (CC BY 4.0).}

\conference{FIRE'21: Forum for Information Retrieval Evaluation,
  December 13--17, 2021, India}

\title{A Feature Extraction based Model for Hate Speech Identification}


\author[1,2]{Salar Mohtaj}[%
email=salar.mohtaj@tu-berlin.de,
url=https://salarmohtaj.github.io/]
\address[1]{Quality and Usability Lab, Technische Universität Berlin, Berlin, Germany}
\address[2]{German Research Centre for Artificial Intelligence (DFKI), Projektbüro Berlin, Berlin, Germany}

\author[1]{Vera Schmitt}[%
email=vera.schmitt@tu-berlin.de,
url= https://vera-schmitt.netlify.app/]

\author[1,2]{Sebastian Möller}[%
email=sebastian.moeller@tu-berlin.de,
url=https://www.qu.tu-berlin.de/menue/team/professur/]

\begin{abstract}
  The detection of hate speech online has become an important task, as offensive language such as hurtful, obscene and insulting content can harm marginalized people or groups. This paper presents TU Berlin team experiments and results on the task 1A and 1B of the shared task on hate speech and offensive content identification in Indo-European languages 2021. The success of different Natural Language Processing models is evaluated for the respective subtasks throughout the competition. 
  We tested different models based on recurrent neural networks in word and character levels and transfer learning approaches based on Bert on the provided dataset by the competition. Among the tested models that have been used for the experiments, the transfer learning-based models achieved the best results in both subtasks.

\end{abstract}


\maketitle

\section{Introduction}
Using abusive language and hate speech in social media platforms can have devastating effects on internet users by promoting racism, hatred and violence \cite{kim2020intersectional}. Offensive language has even the potential to shape political campaigns \cite{gagliardone2016mechachal}. Due to the openness, anonymity and informal structure, social medial platforms are particularly vulnerable to ill-intentioned activities \cite{alatawi2021detecting}. The availability of large annotated corpora from social medial platforms and the development of powerful Natural Language Processing (NLP) has the potential to remedy the challenge of detecting hate speech online \cite{florio2020time}.

Most of the research in this domain is dedicated to English datasets only. Therefore, the Hate Speech and Offensive Content Identification (HASOC) track aims to provide a platform to develop and optimize algorithms for the hate speech detection task in different languages, such as Hindi, German and English \cite{mandl2020overview}. This year HASOC provides a data challenge for multilingual research on the identification of offensive speech online at the Forum for Information Retrieval Evaluation (FIRE) 2021. HASOC has defined two subtasks, whereas the first subtask contains the identification and discrimination of hate, profane and offensive posts from Twitter in English, Hindi and Marathi. The second subtask focuses on the identification of conversational hate-speech in Code-Mixed Languages. 
The TU Berlin team focuses on the subtasks 1A and 1B. Subtask 1A is a coarse-grained binary classification task where tweets should be classified into two classes: 
\begin{itemize}
    \item \emph{\textbf{(NOT)} Non Hate-Offensive: These posts do not contain any hate speech, profane or offensive content}
    \item \emph{\textbf{(HOF)} Hate and Offensive: These posts contain hate, offensive and profane content}
\end{itemize}

Subtask 1B is is a three-class classification task offered for English and Hindi, where hate-speech, profane and offensive posts from subtask 1A are further classified into the following categories: 

\begin{itemize}
\item \emph{\textbf{(HATE} Hate speech: this class contains posts which hate-speech content}
\item \emph{\textbf{(OFFN} Offensive: posts in this class contain offensive content }
\item \emph{\textbf{(PRFN} Profane: posts in this class contain profane content}
\end{itemize}

In this paper the proposed models for classifying tweets into one of the classes for the respective subtask are presented. For this purpose, the state-of-the-art NLP methods are applied to classify the posts and categorize them into the classes. Hereby the team of TU Berlin focuses on the English dataset. We used transfer learning models based on the BERT language model \cite{devlin2018bert}, and also Recurrent Neural Networks (RNNs), either in word and character levels to categorize tweets into the relevant classes. \par

The following section \ref{relatedwork} describes some of the state-of-the-art models for the task of hate speech detection in English. Section \ref{data} describes the provided train and test dataset, whereas section \ref{exp} contains details about data processing and the experiments and models applied. Furthermore, in section \ref{results} the achieved results are analyzed, and section \ref{concl} summarizes and concludes the approaches and results.

\section{Related Work}
\label{relatedwork}
In this section, we overview some of the recent approaches for automatic hate speech detection from English text. Although the automated approaches for hate speech detection could be categorized into keyword-based, source metadata and machine learning based approaches \cite{macavaney2019hate}, in this section we focus on some of the state-of-the-art machine learning based models. \par
Among the proposed models for the HASOC shared task on 2020, \citeauthor{DBLP:conf/fire/MishraSK20} has been used a 
Long Short-Term Memory (LSTM) based model using Glove vectors \cite{DBLP:conf/emnlp/PenningtonSM14} for the embedding \cite{DBLP:conf/fire/MishraSK20}. They fed the outputs of the embedding layer to a single layer LSTM network and put a fully connected layer on top. In this year's competition we tried to use a similar architecture as one of our experiments. On the other side, the \textit{YNU\_OXZ} team \cite{DBLP:conf/fire/OuL20} in the HASOC 2020 competition proposed a model based on \textit{XLM-RoBERTa} \cite{DBLP:conf/acl/ConneauKGCWGGOZ20} and LSTMs. In their model, they concatenated the output of the last layer hidden state of \textit{XLM-RoBERTa} and the hidden state of the last four layers of \textit{XLM-RoBERTa} that is fed into an Ordered Neurons LSTM (ON-LSTM) \cite{DBLP:conf/iclr/ShenTSC19}. Finally, they input these vectors into a fully connected network for the final classification. \par
\citeauthor{DBLP:conf/www/BadjatiyaG0V17} did different experiments based on three different neural network architecture to detect hate speech tweets in Twitter \cite{DBLP:conf/www/BadjatiyaG0V17}. They used convolutional Neural Networks (CNNs), LSTM, and FastText \cite{DBLP:conf/eacl/GraveMJB17}, with either random embeddings or GloVe embeddings.The proposed models categorize tweets as racist, sexist or neither. Their experiments show that the model based on LSTM, random embedding and Gradient
Boosted Decision Trees outperforms the other models in terms of precision, recall, and F1 score.

\section{Data}
\label{data}
The English dataset of HASOC 2021 for the subtasks 1A and 1B, contains the text content of tweets in English, IDs, and the labels for subtask 1A and 1B, respectively. the statistics of the training dataset is presented in Table \ref{tab:datastatistics}. Moreover, the test dataset contains \textbf{1281} tweets which should be categorized into one of the classes based on the subtask. \par 
\begin{table}[b]
\centering
\begin{tabular}{l|c|cc|cccc}
\toprule
\multirow{2}{*}{\textbf{Language}} & \multirow{2}{*}{\textbf{Total \# of Instances}} & \multicolumn{2}{c|}{\textbf{Subtask 1A}} & \multicolumn{4}{c}{\textbf{Subtask 1B}} \\
& & \textbf{HOF} & \textbf{NOT} & \textbf{HATE} & \textbf{OFFN} & \textbf{PRFN} & \textbf{NONE} \\
\midrule
\textbf{English} & 3843 & 2501 & 1342 & 683 & 622 & 1196 & 1342 \\
\bottomrule
\end{tabular}
\caption{\label{tab:datastatistics} Statistics of the \textit{HASOC2021} \textbf{training} dataset for subtasks 1A and 1B}
\end{table}


The content would contains hashtags, emojis, links and usernames that refer to a user on Twitter. A sample of the dataset in different categories is presented in Table \ref{tab:datasamples}. More details about the datasets are provided in \cite{hasoc2021mergeoverview,hasoc2021overview}. \par 

\begin{table}
\centering
\begin{tabular}{p{9cm}p{2cm}p{2cm}}
\toprule
\multirow{2}{*}{\textbf{Sample Tweets}} & \multicolumn{2}{c}{\textbf{Classes}} \\
\cline{2-3}
& \textbf{Sub-task 1A} & \textbf{Sub-task 1B} \\
\midrule
This is enough of yours Modi This is not skill India it is kill India @narendramodi \#ExitModi \#Resign\_PM\_Modi https://t.co/m9FZyU4Lfg & HOF & OFFN \\
\hline
Please, abdicate! You failed us. You failed everyone. Everyone is suffering. EVERYONE! \#ModiKaVaccineJumla &  HOF & HATE \\
\hline
 @Feisty\_Waters Ok. What did you do to piss off the universe?  & HOF & PRFN \\
\hline
@ndtv Nothing gonna help you please \#Resign\_PM\_Modi & NOT & NONE \\
\bottomrule
\end{tabular}
\caption{\label{tab:datasamples} Samples of tweets from the English train dataset in different classes}
\end{table}

\section{Experiments}
\label{exp}
This section contains a short description on the used pre-processing steps and also the developed models and experiments for the task of hate speech detection. 


\subsection{Data Processing}
For pre-processing of the raw data, we followed the same procedure as the experiments on the last year's competition \cite{Mohtaj2020TUBAH}. The data pre-processing mainly includes the replacement of mentions with the phrase \textit{'username'}, replacement of emojis with short textual descriptions, links are also replaced with the phrase \textit{'link'}, and the replacement of multiple white spaces with a single white space. These steps are applied to both, the train and test datasets in order to facilitate the training process.

\subsection{Models}
The best performance of the last year HASOC competition for the English dataset have been achieved by \cite{ankit2020} with a LSTM using GloVe embeddings \cite{DBLP:conf/emnlp/PenningtonSM14} as input. Furthermore, transformer based language models such as BERT \cite{devlin2018bert}, DistilBERT and RoBERTa \cite{Kumar2020ComMAFIRE2E}, and also ELMO \cite{peters2018deep} showed also promising results for similar task. Therefore, the TU Berlin team focuses on BERT based transfer learning approaches for the proposed subtasks. We also did some experiments on character level LSTM models which achieved our best results on the last year's competition \cite{Mohtaj2020TUBAH}. \par

\subsection{LSTM based models}
We developed two different models based on LSTM networks. We developed a smaller, character based architecture, Char\_LSTM hereinafter, and a deeper and more complex network based on words, Word\_LSTM hereinafter. Since people sometimes do minor changes on the words (e.g., by repeating some characters) when they express hate speech, a word based model may not signal those terms properly. As a result, we also developed a character based model to compare the outcomes of the models. \par
For the Char\_LSTM, we tried out different hyper-parameters that includes:
\begin{itemize}
    \item Embedding dimension [50, 100, 200]
    \item Hidden dimension [16, 32, 64, 128]
    \item Dropout [0.25, 0.5, 0.75]
\end{itemize}
The range of the above mentioned hyper-parameters for the Word\_LSTM model are as follow:
\begin{itemize}
    \item Embedding dimension [100, 300]
    \item Hidden dimension [32, 64, 128, 256, 512]
    \item Dropout [0.25, 0.5, 0.75]
\end{itemize}
In our experiments, the batch size of 32 and, the Adam optimizer \cite{DBLP:journals/corr/KingmaB14} and the Binary Cross Entropy (BCE) loss function have been used in both models. In the Word\_LSTM model, we tested either using Glove pre-trained vectors and training the embedding layer from scratch. The detailed results of the proposed models are presented in the section \ref{results}.

\subsection{Bert based models}
In addition to the models based on Recurrent Neural Networks (RNNs), we tested two transfer learning based models using BERT language model \cite{DBLP:journals/corr/abs-1810-04805}. In one of the experiments, we fine-tuned English Bert for the task of hate speech identification. For this purpose, we followed the recommended hyper-parameters by the authors \cite{DBLP:journals/corr/abs-1810-04805}. \par
As the other transfer learning based model, we used Bert for extracting features from textual data. In other words, in this approach, the Bert language model was used to convert text data into vectors. The resulting vectors inputted into a Gated Recurrent Units (GRU) network. Different hyper-parameters tested on the data to choose the best parameters. The range of different hyper-parameters which had been used in the feature extraction approach are as follow:

\begin{itemize}
    \item Hidden dimension [32, 64, 128, 256, 512]
    \item Dropout [0.25, 0.5, 0.75]
\end{itemize}
Like the LSTM based models, the batch size of 32 and, the Adam optimizer \cite{DBLP:journals/corr/KingmaB14} and the Binary Cross Entropy (BCE) loss function have been used in this experiment. We present the detailed results by the different architectures in section \ref{results}.

\section{Results}
\label{results}
In this section the achieved results on the training data are presented. 
For doing the experiments, the training dataset has been divided into train, validation and test datasets. The train part contains 70\% of the whole data, the validation part consist of 10\% of the data, and the test part contains the remaining 20\% of the provided dataset. \par
We tested all of the mentioned models with different hyper-parameters. The best achieved results are shown in tables \ref{tab:charwordlstmresults} - \ref{tab:bertffinetune}. In order to determine the impact of the pre-processing steps on the final results, we've repeated the experiments with the same hyper-parameters without applying the pre-processing steps. Although the runs without applying pre-processing could achieve competitive results in some cases, the experiments based on the pre-processed data outperforms the other ones in most of the cases.
The performance of the submitted models for both sub-tasks are reported in details in \cite{hasoc2021mergeoverview}. 

\begin{table}
\centering
\begin{tabular}{lccccr}
\toprule
\multirow{2}{*}{\textbf{Model name}} & \multirow{2}{*}{\textbf{Pre-processed}} & \multicolumn{3}{c}{\textbf{Hyper-parameters}} & \multirow{2}{*}{\textbf{F1}} \\
\cline{3-5}
& & \textbf{Embedding dimension} & \textbf{Hidden dimension} & \textbf{dropout} & \\
\midrule
 \multirow{6}{*}{\textbf{Char\_LSTM}} & yes & 50 & 256 & 0.5 & 0.75 \\
  & yes & 50 & 128 & 0.75 & 0.78 \\
  & yes & 100 & 64 & 0.5 & 0.76 \\
  & yes & 200 & 16 & 0.5 & \textbf{0.79} \\
  & no & 200 & 16 & 0.75 & 0.75 \\
  & no & 100 & 128 & 0.75 & 0.77 \\
  \midrule
  \multirow{4}{*}{\textbf{Word\_LSTM}} & yes & 100 & 512 & 0.25 & 0.81 \\
  & yes & 300 & 256 & 0.25 & \textbf{0.83} \\
  & yes & 300 & 256 & 0.75 & 0.80 \\
  & no & 300 & 256 & 0.25 & 0.79 \\

\bottomrule
\end{tabular}
\caption{\label{tab:charwordlstmresults} The achieved results by the \textbf{character based} and \textbf{word based LSTM} models for the sub-task 1A}
\end{table}

\begin{table}
\centering
\begin{tabular}{lccccr}
\toprule
\multirow{2}{*}{\textbf{Model name}} & \multirow{2}{*}{\textbf{Pre-processed}} & \multicolumn{3}{c}{\textbf{Hyper-parameters}} & \multirow{2}{*}{\textbf{F1}} \\
\cline{3-5}
& & \textbf{Bert model} & \textbf{Hidden dimension} & \textbf{dropout} & \\
\midrule
 \multirow{5}{*}{\textbf{BERT feature extraction}} & yes & base & 256 & 0.25 & \textbf{0.86} \\
 & yes & base & 128 & 0.25 & 0.83 \\
 & yes & large & 256 & 0.5 & 0.84 \\
 & yes & large & 128 & 0.25 & 0.80 \\
 & no & base & 128 & 0.25 & 0.79 \\
\bottomrule
\end{tabular}
\caption{\label{tab:bertffeatureresults} The achieved results by the \textbf{BERT feature extraction based} model for the sub-task 1A}
\end{table}

\begin{table}
\centering
\begin{tabular}{lccr}
\toprule
\multirow{2}{*}{\textbf{Model name}} & \multirow{2}{*}{\textbf{Pre-processed}} & \textbf{Hyper-parameters} & \multirow{2}{*}{\textbf{F1}} \\
\cline{3-3}
& & \textbf{Bert model}  & \\
\midrule
 \multirow{2}{*}{\textbf{BERT fine-tuning}} & yes & base & 0.81 \\
 & no & base & \textbf{0.83} \\
\bottomrule
\end{tabular}
\caption{\label{tab:bertffinetune} The achieved results by the \textbf{BERT fine-tuning} model for the sub-task 1A}
\end{table}

The same architectures have been trained on the data for the sub-task 1B. The best achieved results on the second task were applied on the sub-task 1B test dataset and submitted to the shared task.

\section{Conclusion and Future Work}
\label{concl}
In this paper, we presented the proposed models on the task 1A and 1B of the shared task on hate speech and offensive content identification in English. We used a BERT based architecture and word and character based LSTM models for training a model to classify tweets into offensive and not offensive categories. Our experiments show that Bert based model outperform the other approaches. \par
Since over-fitting was one of the main issues for training different models during the competition, enriching the training data by adding data samples from different resources could be a possible solution for improving the results. Moreover, the proposed transfer learning based results could be compared with the results from the the other state-of-the-art language models like GPT-3 to check if there is a significant difference in the performances. 

\begin{acknowledgments}
We would like to thank the organizers of \textit{HASOC2021} shared task for organizing the competition and taking time on the inquiries. 
\end{acknowledgments}
\bibliography{sample-ceur}
\end{document}